\documentclass[runningheads]{llncs}
\usepackage{graphicx}
\usepackage{subfigure}
\usepackage{amsmath,amssymb,amsfonts}
\usepackage{color}
\usepackage{ulem}
\usepackage{multirow}
\usepackage{bm}
\usepackage{hyperref}
\hypersetup{
    colorlinks=true,
    linkcolor=blue,
    filecolor=magenta,      
    urlcolor=cyan,
    pdftitle={Overleaf Example},
    pdfpagemode=FullScreen,
    }
\begin{document}
\title{Few-shot Generation of Personalized Neural Surrogates for Cardiac Simulation \\via Bayesian Meta-Learning}
\titlerunning{Few-shot Generation of PNS for Cardiac Simulation}
%
\author{Xiajun Jiang\inst{1} \and
Zhiyuan Li \inst{1} \and
Ryan Missel \inst{1} \and
Md Shakil Zaman \inst{1} \and
Brian Zenger \inst{2} \and
Wilson W. Good \inst{2} \and
Rob S. MacLeod \inst{2} \and
John L. Sapp \inst{3} \and
Linwei Wang\inst{1}}
%
\authorrunning{X. Jiang et al.}
%
\institute{Rochester Institute of Technology, Rochester, NY 14623, USA \\
\email{\{xj7056, zl7904, rxm7244, mz1482, linwei.wang\}@rit.edu} \and
The University of Utah, Salt Lake City, UT 84112, USA \\
\email{brian.zenger@hsc.utah.edu, wilsonwgood@gmail.com, macleod@sci.utah.edu} \and
Dalhousie University, Halifax, NS, Canada \\
\email{john.sapp@nshealth.ca}}
%
\maketitle              
\begin{abstract}
Clinical adoption of personalized virtual heart simulations faces challenges in model personalization and expensive computation. While an ideal solution is an efficient neural surrogate that at the same time is personalized to an individual subject, the state-of-the-art is either concerned with personalizing an expensive simulation model, or learning an efficient yet generic surrogate. This paper presents a completely new concept to achieve \textit{personalized neural surrogates} in a single coherent framework of meta-learning (metaPNS). Instead of learning a single neural surrogate, we pursue the process of learning a personalized neural surrogate using a small amount of context data from a subject, in a novel formulation of few-shot generative modeling underpinned by: 1) a set-conditioned neural surrogate for cardiac simulation that, conditioned on subject-specific context data, learns to generate query simulations not included in the context set, and 2) a meta-model of amortized variational inference that learns to condition the neural surrogate via simple feed-forward embedding of context data. As test time, metaPNS delivers a personalized neural surrogate by fast feed-forward embedding of a small and flexible number of data available from an individual, achieving -- for the first time -- personalization and surrogate construction for expensive simulations in one end-to-end learning framework. Synthetic and real-data experiments demonstrated that metaPNS was able to improve personalization and predictive accuracy in comparison to conventionally-optimized cardiac simulation models, at a fraction of computation.

\end{abstract}
\section{Introduction}
Personalized virtual hearts, customized to the observational data of individual subjects,
have shown promise in clinical tasks such as treatment planning \cite{prakosa2018personalized} and risk stratification \cite{arevalo2016arrhythmia}. Their wider clinical adoption, however, is hindered by two major bottlenecks. First, it remains challenging and time-consuming to calibrate these simulation models to an individual's physiology (\textit{i.e.}, personalization), especially for model parameters that are not directly observable (\textit{e.g.}, material properties) \cite{niederer2020creation}. Second, it is not yet possible to run these simulations at scale due to their high computational cost: this prevents comprehensive testing in clinical pipelines, or rigorous assessment of simulation uncertainties \cite{niederer2020creation}. 

Significant progress has been made in personalizing the parameters of a cardiac model \cite{miller2021implementation,dhamala2018quantifying,wong2015velocity,sermesant2012patient,zettinig2013fast}. Earlier methods typically rely on iterative optimization processes involving repeated calls to the expensive simulation model \cite{wong2015velocity,sermesant2012patient}. Increasing recent works start to leverage modern machine learning (ML) methods such as active learning \cite{dhamala2020embedding}, reinforcement learning \cite{neumann2020machine}, and ML of the input-output relationship between the parameter of interest and available measurements \cite{coveney2021bayesian,giffard2016noninvasive}. While the optimization process is being accelerated with these recent advances, model personalization remains a non-trivial and time-consuming process. Furthermore, the outcome of personalization is still concerned with a simulation model too expensive for clinical adoption at scale.

In parallel, advances in deep learning (DL) have led to a surge of interests in developing efficient neural approximations of expensive scientific simulations \cite{Kasim20published}. Progress in building neural surrogates for cardiac electrophysiology simulations, however, has been relatively limited: initial successes have been mainly demonstrated in 2D settings \cite{cantwell2019rethinking,kashtanova2021ep} with a recent work reporting 3D results on the left atrium \cite{fresca2021pod}. A significant challenge arises from the dependence of these simulations on various model parameters such as material properties, denoted here as $\theta$. Most existing works attempt to learn a single neural function $f(\theta)$ as the surrogate of a simulation model $\mathcal{M}(\theta)$. This raises two challenges. First, to learn an accurate $f(\theta)$ requires a significant amount of training data pairs of $\{ \theta_i, \mathcal{M}(\theta_i) \}$ simulated across the input space of $\theta$: this is computationally challenging, and has not been demonstrated possible in existing works. Second, the resulting neural surrogate is generic and requires the knowledge of a proper patient-specific  $\theta$ -- which is often unknown -- in order to become personalized. 

In summary, ideally we would like in the clinical workflow an efficient simulation surrogate that at the same is personalized to an individual subject. The state-of-the-art, however, is either concerned with optimizing a personalized but expensive simulation model, or learning an efficient yet generic surrogate. One may consider naively combining existing works by first building an accurate neural surrogate $f(\theta)$, and then having it optimized to a subject. This however has not been demonstrated feasible, especially considering the challenge of learning a $f(\theta)$ accurate across the space of $\theta$. Even if feasible, it is a solution that combines two disconnected processes to achieve an otherwise intertwined objective. 

In this paper, we present a completely new concept to achieve \textit{personalized neural surrogates} in a single coherent framework of meta-learning (metaPNS). Our guiding principle is that we are interested in a set of, rather then one single, neural functions as the simulation surrogate: therefore, instead of \textit{learning a single neural surrogate,} we would \textit{learn the process of learning} a personalized neural surrogate from a small number of available data of a subject (\textit{context} data). We cast this in a novel formulation of few-shot generative modeling via Bayesian meta-learning. It has two main elements: 1) a set-conditioned generative model as the neural surrogate for cardiac simulation that, conditioned on the \textit{context} data of an individual, learns to generate \textit{target} simulations not included in the context set, and 2) a meta-model of amortized variational inference (VI) that learns to condition (\textit{i.e.,} personalize) the generative model via feed-forward embedding of context data of \textit{variable} sizes. Compared to optimization-based meta-learning methods \cite{finn2017model,ravi2016optimization}, this type of feed-forward meta-models remove the need of further training and obtain a model at meta-test time via simple feed-forward embedding of context data. With this, at test time, a personalized neural surrogate can be quickly obtained by simple feed-forward embedding of a small and flexible number of data available from a subject. This 1) replaces expensive personalizatoin with fast feed-forward meta-embedding, and 2) delivers a personalized neural surrogate for efficient predictive simulations. To our knowledge, this is the first time  personalization and surrogate constructions are achieved in an end-to-end framework of few-shot generative modeling.

We evaluated metaPNS in synthetic and real-data experiments, in comparison to 1) personalized cardiac simulation models obtained via published optimization methods \cite{dhamala2018high} and 2) a conditional generative model similar to what is presented but  without set conditioning or meta-inference. We demonstrated that metaPNS was able to deliver improved personalization and predictive performance at a fraction of computation cost of conventionally-optimized simulation models. Furthermore, we showed that the presented meta-learning elements are critical to the predictive capacity of the resulting neural surrogates. 

\section{Methodology}

\begin{figure}[!t]
\includegraphics[width=.8\textwidth]{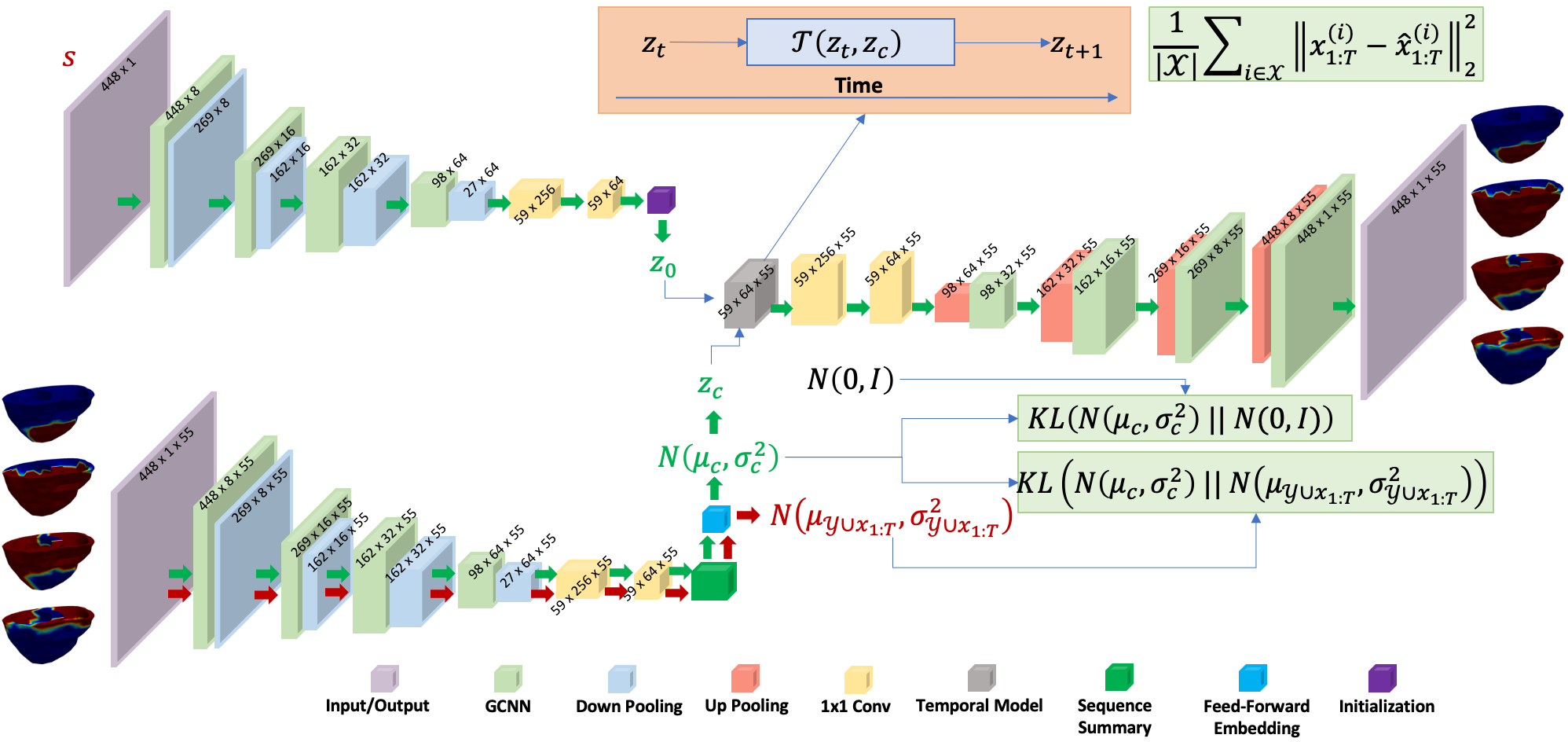}
\caption{Overview of framework. Green arrows show the flow of meta-inference using context set, and red arrows show how target set participates in meta-training.}
\label{overview}
\end{figure}

Fig.~\ref{overview} gives an overview of metaPNS. Consider a cardiac electrophysiology simulation model $x_{1:T} = \mathcal{M}(s;\theta)$ with both known input $s$ and unknown parameter $\theta$: in predictive tasks, a typical example of $s$ is the electrical stimulation applied to the virtual heart, whereas a typical example of $\theta$ is the material property to be tailored to an individual. Consider observations $y_{1:T}$ of $x_{1:T}$ that, for instance, can be sparse heart surface voltage mapping, or body-surface potential recordings. The goal of model personalization is to obtain an estimated $\hat{\theta}$ to minimize the actual and simulated observations. The goal of personalized neural surrogate is then to obtain a neural function $f(s,\hat{\theta})$ that 1) can approximate and accelerate the computation of $\mathcal{M}(s;\hat{\theta})$ and 2) is calibrated to an individual's observations. 

It is natural to approach this by learning a neural function $f(s, \theta)$ that approximates $\mathcal{M}(s;\theta)$ across the space of $\theta$ \cite{kashtanova2021ep,fresca2021pod}---a generic surrogate that, once learned, requires an input of $\theta$ or a separate optimization process to become personalized. Departing from the common practice, we will aim to obtain a set of personalized neural surrogates that change with a small set of context observations, $\mathcal{Y} = \{ y_{1:T}^{(i)} \}_{i=1}^{\nu}$ with variable size $\nu$, from an individual subject:
\begin{equation}
    \begin{aligned}
    p(\hat{x}_{1:T} | s, \mathcal{Y}) &= \int p(\hat{x}_{1:T} |s,\mathbf{c}) q_{\zeta}(\mathbf{c} | \mathcal{Y} ) dc,
    \end{aligned}
\end{equation}
where the neural surrogate $ p(\hat{x}_{1:T} | s, \mathcal{Y})$ is a stochastic process learned in the context of Bayesian meta-learning: $p(\hat{x}_{1:T} | s,\mathbf{c})$ is a generative model conditioned on known $s$ and personalized by latent embedding $\mathbf{c}$ derived from subject-specific context data; $q_{\zeta}(\mathbf{c} | \mathcal{Y})$ is the meta-model that, parameterized by $\zeta$, learns to personalize the generative model via feed-forward embedding of $\mathcal{Y}$.

\subsection{Set-Conditioned Generative Model}

The generative neural surrogate 
is conditioned on context-set embedding $p(\mathbf{c})$ and known stimulation $s$. It has a temporal transition model $\mathcal{T}$ of the latent state $z_t$, and a spatial model $\mathcal{G}$ for emission to $x_t$ at the high-dimensional cardiac mesh: 
\begin{equation}
    \begin{aligned}
    \mathrm{Transition:}\, &z_{t+1} = \mathcal{T}(z_t), \\
    \mathrm{Emission:}\, &\hat{x}_{t+1} = \mathcal{G}(z_{t+1}).
    \end{aligned}
\end{equation}
where the initial state $z_0$ is obtained by applying a neural function consisting a linear layer to the embedding of the known stimulation input $s$: $z_0 = f_{\rho}(s)$. 

\subsubsection{Spatial Modeling via Graph CNNs (GCNNs):}
As $x$ lives on 3D geometry of the heart, we describe $\mathcal{G}$ with GCNNs. We represent triangular meshes of the heart as undirected graphs, with edge attributes between vertices as normalized differences in their 3D coordinates if an edge exists. Encoding and decoding are performed over hierarchical graph representations of the heart geometry, obtained by a specialized mesh coarsening method \cite{cgal:c-tsms-07}. A continuous spline kernel for spatial convolution is used such that it can be applied across graphs \cite{fey2018splinecnn}. To make the network deeper and more expressive, we introduce residual blocks here 
through a skip connection with 1D convolution \cite{jiang2020learning}.

\subsubsection{Temporal Modeling via Set-Conditioned Transition Function:}
The temporal transition function $\mathcal{T}$ is inspired by Gated Recurrent Units (GRUs) \cite{chung2014empirical}.
Note that, with regular GRUs, $\mathcal{T}$ would be global to the training data rather than subject-specific. Instead, we condition $\mathcal{T}$ on 
the context-set embedding by creating conditional gated transition functions: 
\begin{equation}
    \begin{aligned}
    g_t &= \sigma(W_1 z_t^{(1)} + b_1), \quad z_t^{(1)} = \mathrm{ELU}(\alpha_1 z_t + \beta_1 \mathbf{c} + \gamma_1), \\
    h_t &= \mathrm{ELU}(W_2 z_t^{(2)} + b_2), \quad z_t^{(2)} = \mathrm{ELU}(\alpha_2 z_t + \beta_2 \mathbf{c} + \gamma_2) \\
    z_{t+1} &= (1 - g_t) \odot (W_3 z_t^{(3)} + b_3) + g_t \odot h_t, \quad z_t^{(3)} = \alpha_3 z_t + \beta_3 \mathbf{c} + \gamma_3
    \end{aligned}
\end{equation}
where $\mathbf{c}$ is sampled from the context-set embedding, and $\{W_i, b_i, \alpha_i, \beta_i, \gamma_i \}_{i=1}^3$ are learnable parameters. The model has flexibility to choose a linear transition for some dimensions and non-linear transition for the others.

\subsection{Meta-Model for Amortized Variational Inference}

\subsubsection{Amortized Variational Inference:}
Consider a multi-subject dataset $\mathcal{X} = \{\mathcal{X}^k \}_{k=1}^K, k \in \{ 1, 2, ..., K \}$, where $\mathcal{X}^k = \{ x_{1:T}^{(1)}, x_{1:T}^{(2)},..., x_{1:T}^{(N_k)} \}$ are heart signals from subject $k$ with $N_k$ samples. For each subject $k$, a subset of $\mathcal{X}^k$ is associated with observations $\mathcal{Y}^k = \{ y_{1:T}^{(1)}, y_{1:T}^{(2)},..., y_{1:T}^{(M_k)} \}, M_k \ll N_k$, such that $\mathcal{Y}^k$ is the context set. The rest of $\mathcal{X}^k$ is unobserved target set. Stimulation inputs $\bm{s}^k$ are known on all samples in $\mathcal{X}^k$. The evidence lower bound (ELBO) we optimize is: 
\begin{equation}
    \begin{aligned}
    \sum_k \sum_{x_{1:T} \in \mathcal{X}^k} \log p(\hat{x}_{1:T} | \bm{s}^k, \mathcal{Y}^k) &\geq \sum_k \sum_{x_{1:T} \in \mathcal{X}^k} \mathbb{E}_{q_{\zeta}(\mathbf{c}^k | \mathcal{Y}^k)} \left [ \log p(\hat{x}_{1:T} | \mathbf{c}^k, \bm{s}^k) \right] \\
    &- \mathrm{KL} \left( q_{\zeta}(\mathbf{c}^k | \mathcal{Y}^k \cup x_{1:T}) || p(\mathbf{c}^k | \mathcal{Y}^k) \right),
    \end{aligned}
\end{equation}
where we let $p(\mathbf{c}^k | \mathcal{Y}^k)$ and $q_{\zeta}(\mathbf{c}^k | \mathcal{Y}^k \cup x_{1:T})$ share the same meta set-embedding networks to parameterize their means and variances. We further regularize $p(\mathbf{c}^k | \mathcal{Y}^k)$ to be close to a standard Gaussian distribution $\mathcal{N}(0, I)$. Therefore, the overall loss function is:
\begin{equation}
\label{eqn:metaloss}
    \begin{aligned}
    \arg\min_{\theta} \, &\sum_k \sum_{x_{1:T} \in \mathcal{X}^k} \mathbb{E}_{q_{\zeta}(\mathbf{c}^k | \mathcal{Y}^k)} \left [ \log p(\hat{x}_{1:T} | \mathbf{c}^k, \bm{s}^k) \right] \\ &- \lambda_1 \mathrm{KL} \left( q_{\zeta}(\mathbf{c}^k | \mathcal{Y}^k \cup x_{1:T}) || p(\mathbf{c}^k | \mathcal{Y}^k) \right)
    - \lambda_2 \mathrm{KL} \left( 
    p(\mathbf{c}^k| \mathcal{Y}^k) || \mathcal{N}(0, I) \right),
    \end{aligned}
\end{equation}
where $\lambda_1$ and $\lambda_2$ are regularization multipliers. 

\subsubsection{Context-Set Feed-Forward Embedding:}
We use the meta-model $q_{\zeta}(\bm{c} | \mathcal{D}_c)$ to get the feed-forward embedding for each context set. First, each sample $y_{1:T} \in \mathcal{D}_{c}$ is embedded through a neural function $h_{\phi}(y_{1:T})$ that uses a GCN-GRU cell \cite{jiang2021label} to obtain the sequential information from the graph, and aggregate it across time with a linear layer. We then average all latent embedding in $\mathcal{D}_c$:
\begin{equation}
    \begin{aligned}
    \frac{1}{|\mathcal{D}_c|} \sum\nolimits_{y_{1:T} \in \mathcal{D}_c} h_{\phi}(y_{1:T}),
    \end{aligned}
\end{equation}
which then parameterizes $q_{\zeta} = \mathcal{N}(\bm{\mu}_c, \bm{\sigma}_c^2)$ via two separate linear layers. The conditional factor $\mathbf{c}$ is then sampled by $\mathbf{c} = \bm{\mu}_c + \bm{\epsilon} \odot \bm{\sigma}_c$, where $\bm{\epsilon} \sim \mathcal{N}(0, \mathbf{I})$ \cite{kingma2013auto}. 

The loss in Equation \ref{eqn:metaloss} is optimized in episodic training. In each training episode across all subjects, the input data is divided into two separate sets: a context set $\mathcal{D}_c^k$ consists of small sets of samples from each subject and the target set formed by the remaining data. The model is asked to take $\mathcal{D}_c^k$ for each subject $k$, and generate samples in  $\mathcal{D}_x^k$ including both context and target sets.

\section{Experiments}

In all experiments, metaPNS consists of 4 GCNN blocks and 2 regular convolution layers in the encoders, 1 context-set aggregator with a GCN-GRU block followed by a linear layer to compress time and another linear layer for feature extraction, 1 conditional gated transition unit for the generation of latent dynamics, and 4 GCNN blocks and 2 regular convolution layers in the decoder. We used Adam optimizer \cite{kingma2014adam}. The learning rate is set at $1 \times 10^{-3}$ with a learning rate decreasing rate 0.5 every 50 episodes. The two KL multipliers are: $\lambda_1 = 10^{-4}$ and $\lambda_2 = 0.1$. All experiments were run on NVIDIA Tesla T4s with 16 GB memory. Our implementation is available here: \url{https://github.com/john-x-jiang/epnn}.

\begin{table}[!t]
\centering
    \caption{Performance metrics of 1) the presented metaPNS, 2) PNS without meta-model, 3) FS-BO, and 4) VAE-BO on context and target sets.}
    \label{sim}
    \begin{tabular}{ |c|c|c|c|c|c|c|  }
    \hline
     & \multicolumn{3}{|c|}{Context set} & \multicolumn{3}{|c|}{Target set} \\
    \hline
    Model & MSE & CC & DC & MSE & CC & DC\\
    \hline
    metaPNS & 4.5$\pm$1.2e-4 & 0.74$\pm$0.089 & 
    {\fontseries{b}\selectfont 0.88$\pm$0.10}&
    {\fontseries{b}\selectfont 4.4$\pm$1.1e-4} & {\fontseries{b}\selectfont 0.72$\pm$0.092} & {\fontseries{b}\selectfont 0.88$\pm$0.089}\\
    PNS & {\fontseries{b}\selectfont 2.7$\pm$0.54e-4} & {\fontseries{b}\selectfont 0.83$\pm$0.068} & 0.77$\pm$0.13 & 1.2$\pm$0.29e-3 & 0.44$\pm$0.13 & 0.75$\pm$0.13\\
    FS-BO & 5.3$\pm$6.5e-4  & 0.69$\pm$0.25 & 0.48$\pm$0.34 & 5.3$\pm$5.9e-4 & 0.69$\pm$0.24 & 0.48$\pm$0.34\\
    VAE-BO & 4.8$\pm$2.5e-4 & 0.46$\pm$0.15 & 0.48$\pm$0.09 & 5.0$\pm$3.3e-4 & 0.48$\pm$0.11 & 0.48$\pm$0.12\\
    \hline
    \end{tabular}
\end{table}

\subsection{Synthetic Experiments}

\paragraph{Data and Training.} We generated propagation of action potential by the Aliev-Panfilov model \cite{aliev1996simple} and the corresponding sparse heart-surface potential as measurements. We considered 3 heart meshes with a combination of 16 different tissue parameter settings (15 scar and one healthy) and approximately 200 different locations of stimulations. This is treated as 16 unique subjects. In each episode of meta-training, for each subject, we randomly sample 25 origins with a variable number (1-5) of origins as the context and the rest 20 as the target. Each training episode took on average 8.5 minutes.

\begin{figure}[!t]
\includegraphics[width=.9\textwidth]{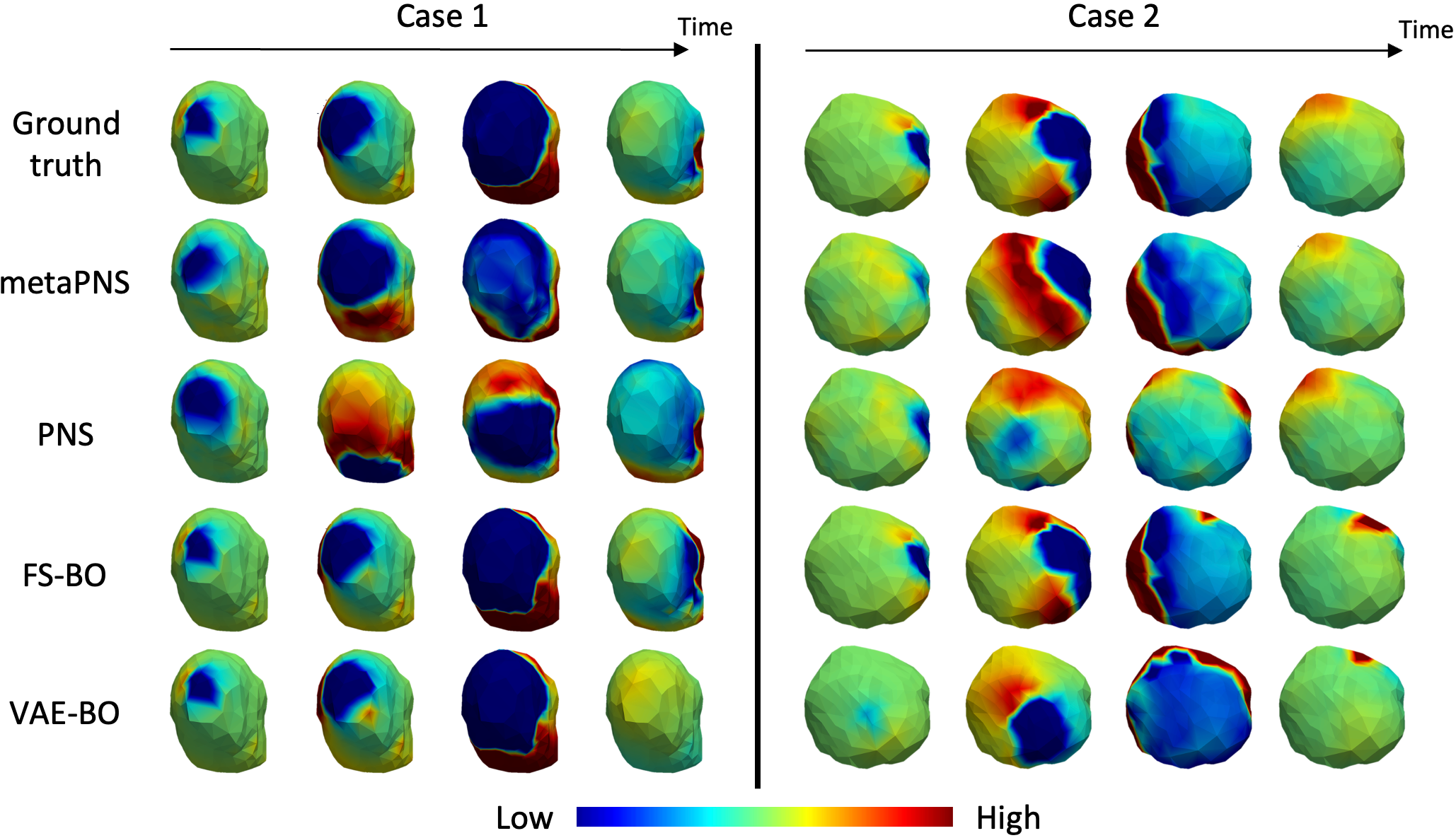}
\caption{Examples of generated simulations on target sets.} \label{sim_fig}
\end{figure}

\paragraph{Testing and Comparisons.} 
In meta-testing, we considered the same 16 subjects with 2,731 unique simulations. For each subject, 3 different context sets with a varying number of samples (1-5) were used to obtain the personalized surrogate, 
to each then simulation 1,638 target samples with distinct stimulations.

The closest related works are those utilizing expensive optimizations to personalize the parameters of a cardiac simulation model, and using the expensive personalized model in predictions. Therefore, for a subset of 11 subjects and 165 unique simulations, we compared metaPNS with published personalized cardiac simulation workflows where \textit{tissue excitability} within the Alieve-Panfilov model was estimated by deriative-free Bayesian optimization \cite{dhamala2018high} using the same context sets used for metaPNS. We considered two optimization formulations where the tissue excitability was represented either using a seven-segment division of the cardiac mesh (FS-BO) or a modern VAE-based generative model as described in \cite{dhamala2018high} (VAE-BO). Predictive simulations were then performed with the personalized cardiac model and compared to metaPNS on the same target samples.

Alternative neural surrogates of cardiac simulations \cite{cantwell2019rethinking,kashtanova2021ep,fresca2021pod} are also related, although existing works are either 2D on image grids or on atria \cite{fresca2021pod}. Once learned, they all have to be separately optimized to a subject’s data for personalized predictions – the latter we have not seen in published works. Thus, for a neural baseline, we considered a version of metaPNS without the set-conditioning or meta-model, \textit{i.e.}, a regular generative model $p(x_{1:T} |s,\mathbf{c})$ where the embedding $c$ is directly inferred from $y_{1:T}$ as $q(\mathbf{c} | y_{1:T})$. We term this PNS. 

\paragraph{Results.} The performance of all methods was quantitatively measured by mean square error (MSE) and spatial correlation coefficient (CC) between the predicted and actual target simulations, and the dice coefficient (DC) of the abnormal tissue region obtained by thresholding signals with Otsu's method \cite{otsu1979threshold}. Performance on the context set shows how well each method fits the data, whereas that on the target set shows how well each personalized model predicts new simulations. As shown in Table.~\ref{sim} and Fig.~\ref{sim_fig}, metaPNS has the strongest predictive performance on target sets. Due to the difficulty in convergence, for FS-BO we set the optimization bound to consider the ground-truth tissue-property setting. Despite this, metaPNS was able to deliver comparable performance in some cases (\textit{e.g.}, Case 1 in Fig.~\ref{sim_fig}), and the highest average accuracy across all samples in target sets. Notably, this performance gain was achieved at a fraction of computation cost: one Alieve-Panfilov simulation took on average 5 minutes versus 0.24 seconds by the neural surrogate; furthermore, an average BO personalization required 100 calls to the simulation model, whereas metaPNS took on average 0.032 seconds for the feed-forward embedding of the context set.

\begin{figure}[!t]
\includegraphics[width=.9\textwidth]{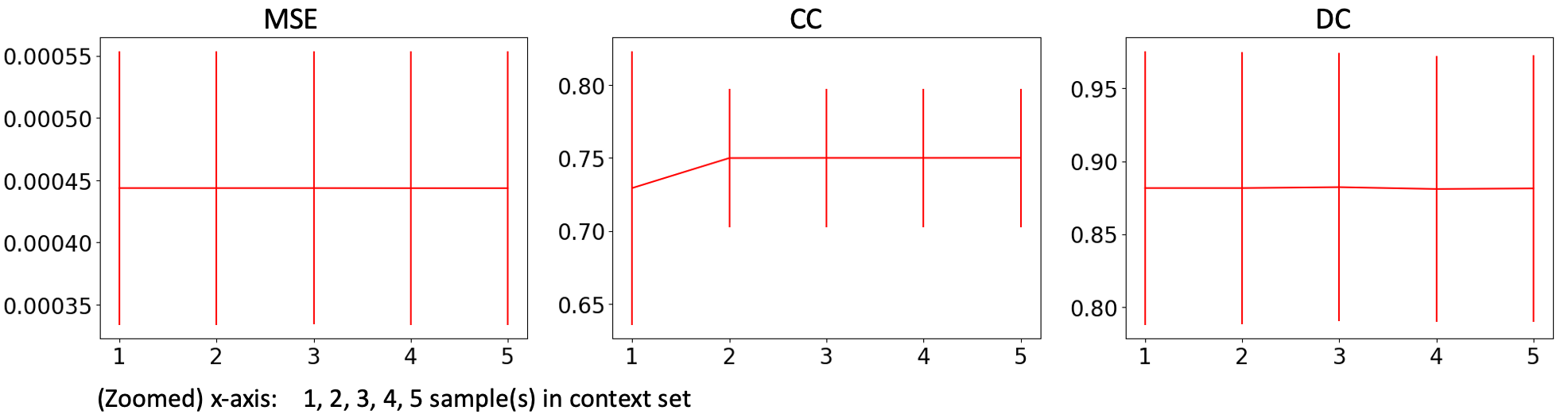}
\caption{Performance of metaPNS as the number of context data decreases.} 
\label{context}
\end{figure}

Compared to metaPNS, PNS was able to well reconstruct the context set (and thus capture the abnormal tissue). It however was not able to generate any new simulations in target sets. This provided evidence for the importance of the presented meta-model of context-set embedding.  
We further compared metaPNS using different number of samples in the context set. As shown in Fig.~\ref{context}, the predictive performance of metaPNS seldomly changed as the context data decreased, demonstrating its strong ability as a few-shot generative model. 

\subsection{Clinical Data}

We then evaluated the trained metaPNS on \textit{in-vivo} recordings from an animal model experiment \cite{RSM:Ber2021}. Several cardiac activation sequences were generated using bipolar stimulation from intramural plunge needles at four sites: left ventricular (LV) base, LV Apex, LV freewall, and LV septum. The cardiac potentials were recorded at the epicardial surface with an epicardial sock with 247 electrodes (inter-electrode spacing 6.5$\pm$1.3~mm). Geometric surfaces were constructed based on electrode locations acquired during each experiment. 

\begin{figure}[!t]
\includegraphics[width=.9\textwidth]{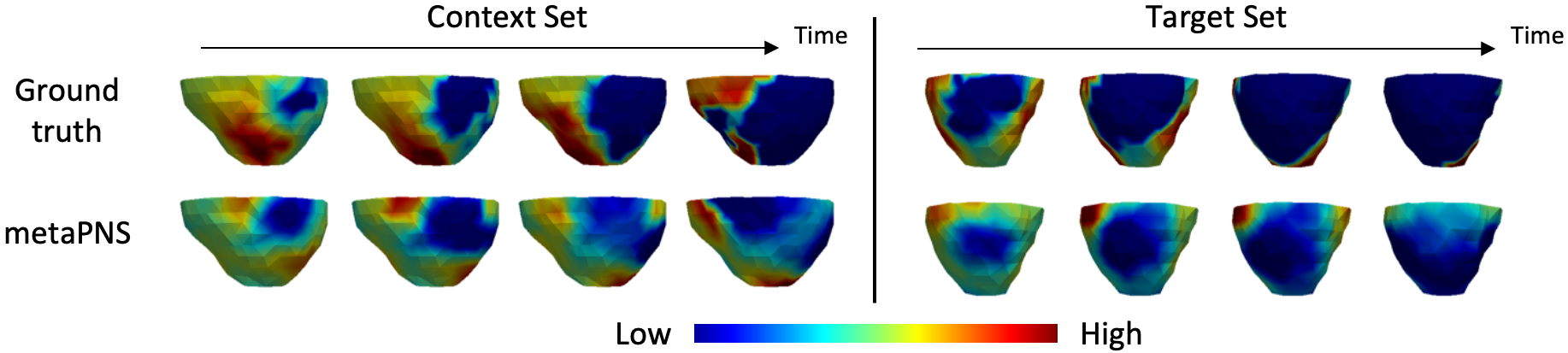}
\caption{Examples of generated target simulations in real-data experiments.} \label{real_fig}
\end{figure}

We carried out cross validation by leaving out one stimulated site as the target sample each time, and the rest as the context set (15 sequences). Accuracy of metaPNS prediction on the target sample was evaluated by MSE and CC with the epicardial sock measurements. Across all samples, metaPNS obtained MSE 1.0$\pm$0.015e-3 and CC 0.36$\pm$0.038 on context sets, MSE 1.0$\pm$0.073e-3 and CC 0.33$\pm$0.087 on target sets. Fig.~\ref{real_fig} show the performance of the presented model on both context and target sets. Despite the performance gap with synthetic experiments, this experiment demonstrated the ability of metaPNS to generalize outside the geometry and simulations seen at the meta-training time.

\section{Conclusion}

In this work, we demonstrated the promising potential of metaPNS -- a novel framework for obtaining a personalized neural surrogate by simple feed-forward embedding of a small and flexible number of data available from a subject. Future work will evaluate use of metaPNS for higher-fidelity cardiac simulations, extend its ability to personalize using different types of observational data, and improve its generalization ability to more complex real-data applications.\\

\noindent\textbf{Acknowledgement:} This work is supported by the National Institutes of Health (NIH) under Award Numbers R01HL145590, R01NR018301, and F30HL149327; the NIH NIGMS Center
for Integrative Biomedical Computing (www.sci.utah.edu/cibc), NIH NIGMS grants P41GM103545
and R24 GM136986; the NSF GRFP; the Utah Graduate Research Fellowship; and the Nora Eccles
Harrison Foundation for Cardiovascular Research.. 

%
%
%
\bibliographystyle{splncs04}
\bibliography{ref}
\end{document}